\documentclass[lettersize,journal]{IEEEtran}
\ifCLASSOPTIONcompsoc
  \usepackage[nocompress]{cite}
\else
  \usepackage{cite}
\fi

\usepackage{xcolor}
\usepackage{multirow}
\usepackage{enumitem}
\usepackage{microtype}
\usepackage{bm}

\usepackage[colorlinks]{hyperref}

\usepackage{booktabs}
\usepackage{amsmath,amsfonts,amssymb,amsthm}
\usepackage[linesnumbered,vlined,boxed,ruled]{algorithm2e}
\usepackage{array}
\usepackage[caption=false,font=footnotesize,hangindent=10pt]{subfig}
\usepackage{graphicx}
\usepackage{tabularx}
\graphicspath{{./}{pics/}}

\usepackage{mdframed}
\newcolumntype{C}[1]{>{\centering\arraybackslash\hspace{0pt}}p{#1}}
\newcolumntype{L}[1]{>{\baselineskip=10pt}m{#1}} 

\usepackage[capitalise]{cleveref}
\crefformat{section}{\S#2#1#3}

\newtheorem{definition}{Definition}
\newtheorem{example}{Example}

\newcommand{\exam}[1]{``\texttt{#1}''}

\newcommand{\bx}{\mathbf{x}}

\usepackage{changes}
\usepackage{lipsum}

\begin{document}

\title{Extracting Training Dialogue Data
from Large Language Model based Task Bots}
\author{
  Shuo~Zhang,
  Junzhou~Zhao,
  Junji~Hou,
  Pinghui~Wang,
  Chenxu Wang,
  Jing~Tao

\IEEEcompsocitemizethanks{
\IEEEcompsocthanksitem
S. Zhang, J. Zhao, J. Hou, P. Wang, C. Wang and J. Tao
are with the MOE Key Laboratory for Intelligent Networks and Network Security,
Xi'an Jiaotong University, P.O. Box 1088,
No. 28, Xianning West Road, Xi'an, Shaanxi 710049, China.
E-mail: \{zs412082986, 15955192\}@stu.xjtu.edu.cn, \{junzhou.zhao, phwang, cxwang, jtao\}@mail.xjtu.edu.cn
(Corresponding author: Junzhou~Zhao.)
}

}



\IEEEtitleabstractindextext{
\begin{abstract}
  Large Language Models (LLMs) have been widely adopted to enhance Task-Oriented
Dialogue Systems (TODS) by modeling complex language patterns and delivering
contextually appropriate responses.
However, this integration introduces significant privacy risks, as LLMs,
functioning as soft knowledge bases that compress extensive training data into
rich knowledge representations, can inadvertently memorize training dialogue
data containing not only identifiable information such as phone numbers but also entire dialogue-level events like complete travel
schedules.
Despite the critical nature of this privacy concern, how LLM memorization is
inherited in developing task bots remains unexplored.
In this work, we address this gap through a systematic quantitative study that
involves evaluating existing training data extraction attacks, analyzing key
characteristics of task-oriented dialogue modeling that render existing methods
ineffective, and proposing novel attack techniques tailored for LLM-based TODS
that enhance both response sampling and membership inference.
Experimental results demonstrate the effectiveness of our proposed data
extraction attack.
Our method can extract thousands of training labels of dialogue states with best-case
precision exceeding $70\%$.
Furthermore, we provide an in-depth analysis of training data memorization in
LLM-based TODS by identifying and quantifying key influencing factors and
discussing targeted mitigation strategies.

\end{abstract}

}

\maketitle

\IEEEdisplaynontitleabstractindextext
\IEEEpeerreviewmaketitle

\section{Introduction} \label{sec:introduction}

\IEEEPARstart{T}{ask}-Oriented Dialogue Systems (TODS), or task bots, play a
vital role in assisting users with specific tasks such as travel bookings and
medical consultations~\cite{algherairy2023review}.
Trained on massive datasets that encompass a broad spectrum of human knowledge,
Large Language Models (LLMs) have substantially enhanced the performance of TODS
by enabling more accurate understanding of user intentions and generate relevant
responses~\cite{kulhanek2021augpt, peng2021soloist, yang2021ubar,
  hosseini2020simple, sun2022mars, he2022galaxy}.
Despite the substantial advancements, the integration of LLMs introduces significant
privacy risks as LLMs possess the capability to memorize and inadvertently leak
sensitive training data~\cite{carlini2021extracting, carlini2022quantifying,
  yu2023bag}.
Training data memorization is commonly attributed to the necessity of
retaining specific information for effective response generalization, wherein
models rely on copying and interpolating from individual training
examples~\cite{carlini2023extracting}.

While training data memorization in LLMs has been studied previously, there
remains a research gap in understanding how such memorization is inherited when
LLMs are finetuned for TODS.
This research gap is particularly concerning given the pervasive deployment of
TODS in real-world applications worldwide.
These systems inherently handle sensitive and personal data, including not only
identifiable information such as phone numbers and personal addresses but also
dialogue-level events that encapsulate detailed user personal schedules and
preferences such as travel plans and hobbies.
Let us consider the following example.

\begin{figure}[t]
  \centering
  \includegraphics[width=.48\textwidth]{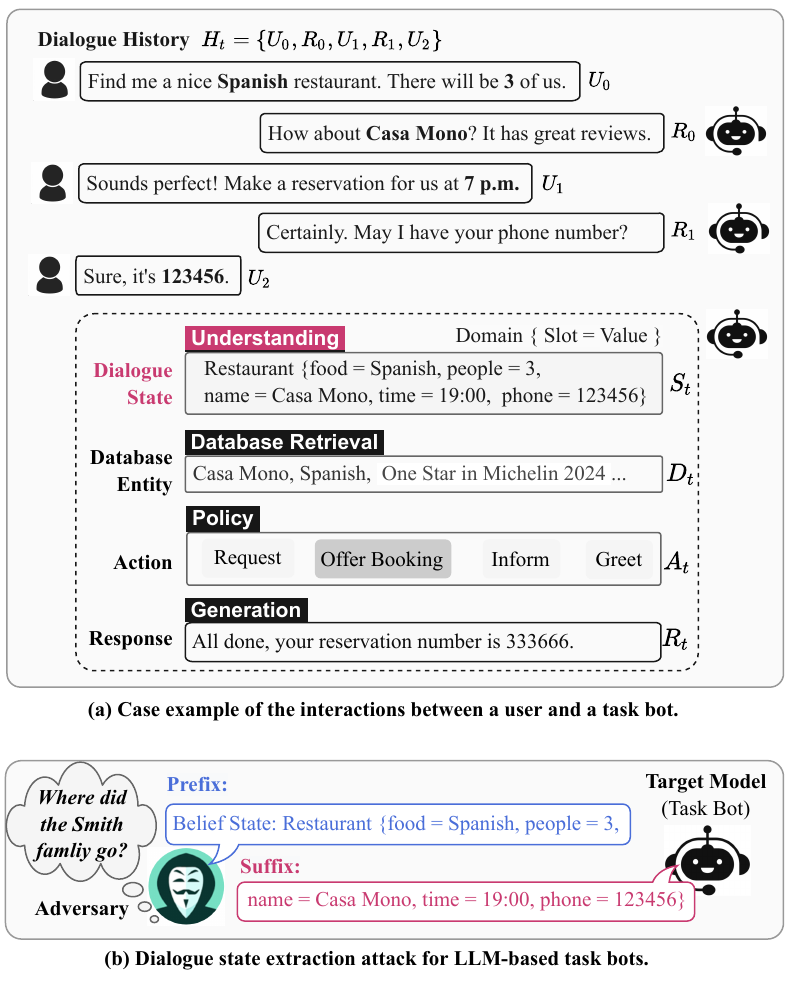}
  \caption{(a) LLM-based task bot utilizes a single neural auto-regressive model to
    parameterize the sequential dialogue pipeline.
    (b) Our dialogue state extraction attack extracts the training
    labels of \textit{\textbf{Dialogue States}} $(S)$ \textit{\textbf{without}} access to the corresponding conditioning text of dialogue
    histories.}
  \label{fig:train_data}
\end{figure}

\IEEEpubidadjcol

\begin{figure*}[t]
  \includegraphics[width=.95\linewidth]{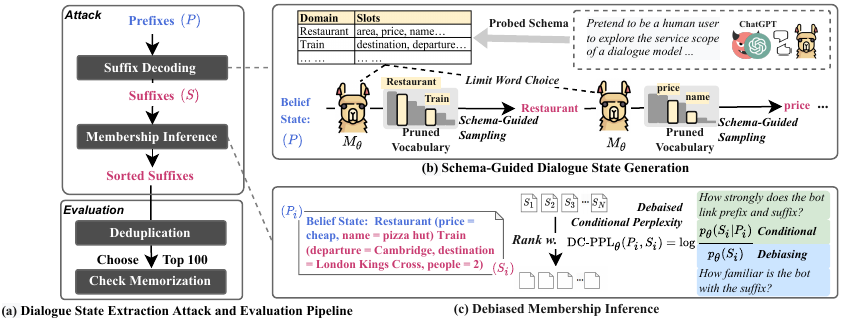}
  \centering
  \caption{\textbf{(a) Workflow of the dialogue state extraction attack:} We begin
    with \textbf{Suffix Decoding}, generating many samples from the LLM-based
    task bot by seeding the model with either empty inputs (for untargeted
    attack) or partial dialogue state prefixes (for targeted attack).
    \textbf{Membership Inference} metrics are then applied to rank each
    generation, with higher ranks indicating a greater likelihood of originating
    from the training data.
    \textbf{For evaluation}, we deduplicate generations, keep the top 100, and
    mark each generation as either memorized or not memorized by comparing them
    with the dialogue states from the training dataset.
    \textbf{(b) Schema-Guided Sampling \cref{sec:lmvslm}} generates more valid
    results by constraining word choices for sampling domains and slots using
    the dialogue schema (i.e., service scope), which is automatically explored
    by simulating user interactions with ChatGPT.
    \textbf{(c) Debiased Conditional Perplexity \cref{sec:adv_mia}} reduces bias
    toward favoring common generic dialogue state components.}
  \label{fig:framework}
\end{figure*}

\begin{example}[Restaurant Booking Assistant Bot]\label{exam:restaurant}
  A restaurant booking assistant bot helps users find restaurants and make
  reservations.
  As illustrated in \cref{fig:train_data}, a dialogue comprises multiple turns of
  user-system utterances.
  At each turn, the bot tracks the \textbf{dialogue state} based on the dialogue
  history.
  The dialogue state represents a collection of user preferences and constraints such as
  food type and booking time, extracted from prior user utterances.
  The bot then queries the database according to the dialogue state, predicts the
  appropriate system actions, e.g., offering a reservation, and generates an
  appropriate response.
  In practice, the dialogue state is a part of the training data.
  Consequently, an adversary with access to the task bot API may maliciously
  induce the bot to output its dialogue state, e.g., \exam{Restaurant(name=Casa
    Mono,food=Spanish,time=19:00,phone=123456, people=3)}, thus leaking a user's
  private information.
\end{example}

In this work, we investigate how LLM-based task bots memorize training dialogue data—specifically, dialogue states that maintain constraints from the dialogue history—by performing training data extraction attacks~\cite{carlini2021extracting, yu2023bag, carlini2023extracting}. These attacks aim to recover training labels from a fine-tuned language model via score-based black-box query access, and typically follow a two-stage pipeline. As illustrated in \cref{fig:framework}, the first stage prompts the model to generate numerous candidate dialogue states through various sampling strategies in a suffix decoding step. The second stage applies membership inference metrics to rank these candidates, flagging those most likely to have originated from the training dataset.

{
Extracting training dialogue states from LLM-based task bots is inherently challenging due to the structural characteristics of TODS.
Unlike open-ended dialogue models, a TODS is never trained to reproduce user input utterances—thus, raw inputs are not memorized and cannot be extracted from the model.
Instead, the model is optimized to predict structured dialogue states conditioned on preceding context, making extraction possible only through the output space.
Furthermore, a TODS must maintain consistency and coherence across multiple dialogue turns~\cite{algherairy2023review}. When the preceding context is partially observed or truncated, recovering the correct subsequent state becomes extremely challenging.
Finally, even when the entire context is known, the one-to-many nature of
dialogue responses (i.e., multiple correct replies can exist for the same
preceding context) further complicates data extraction, as an extracted
response may not necessarily reflect the original training data.
}



To address these challenges, we make two primary contributions. First, we observe that incomplete preceding context typically causes existing suffix decoding methods to generate incoherent or nonsensical dialogue states. To mitigate this, we propose a \emph{schema-guided sampling method} that leverages the dialogue schema (i.e., service scope) to constrain word choices, ensuring diversity and validity in the generated outputs. Second, we find that the one-to-many nature of dialogue leads to model bias in existing membership inference metrics, which tend to favor information-poor, generic dialogue fragments (e.g., greetings). To overcome this, we introduce a \emph{debiased conditional perplexity metric}, which evaluates input–output entailment while mitigating the model’s overfamiliarity with generic fragments.

Experimental results demonstrate that our method effectively extracts hundreds
to thousands of training dialogue states.
We consider two extraction settings, i.e., \emph{targeted extraction} and
\emph{untargeted extraction}. Targeted extraction
prompts the task bot with partial dialogue state, while untargeted extraction
does not.
In the untargeted extraction setting, we find that values in the dialogue states
(e.g., name) exhibit a relatively higher privacy leaking risk, with a maximum
extraction precision reaching $67\%$, while full dialogue states present less
concern, with a maximum precision of $26\%$.
In the targeted extraction setting, our method shows a significant increase in
precision, achieving a best-case scenario of $100\%$ precision for individual
values and over $70\%$ for event-level states, indicating a substantially
greater privacy leaking concern.
Furthermore, we identify critical factors affecting task bot memorization, i.e.,
\emph{increased substring repetition inherent in turn-level dialogue
  training data}, which enhances memorization, and the \emph{one-to-many feature
  of dialogues}, where multiple responses can correspond to a single prompt, which
can reduce memorization.
Based on our findings, we further discuss practical mitigating strategies
including dialogue-level modeling to address substring repetition and the value
copy mechanism to mitigate contextual correlations.

Our contributions are summarized as follows:
\begin{itemize}

\item To the best of our knowledge, this is the first work to investigate the
  training data memorization issue in LLM-based task bots, where privacy leakage emerges at the level of structured belief states and cross-slot task semantics rather than isolated utterance fragments.

\item We present the first systematic investigation including  {(i) framing the TODS data extraction threat model, (ii) systematically benchmarking existing techniques in this new context, (iii) identifying domain-specific biases and challenges, and (iv) proposing two targeted methodological adaptations tailored for this task: a schema-guided sampling strategy that efficiently generates valid dialogue state candidates by leveraging task schema constraints, and a debiased conditional perplexity metric that corrects for schema frequency bias in existing perplexity-based membership inference approaches.}

\item Extensive experiments across various configurations demonstrate the severe
  and practical implications of our attack techniques, highlighting substantial
  privacy leaking risks.
  We identify factors influencing memorization in LLM-based task bots and
  discuss targeted mitigation strategies to guide future implementations.
\end{itemize}


\section{Preliminaries and Related Work}
\label{sec:formluation}

In this section,
we first elaborate some background on LLM-based task bots and the finetuning process, then we introduce existing training data extraction attacks.

\subsection{Large Language Model based Task Bots}
Task-oriented dialog systems, or task bots, are designed to help users complete
specific tasks through multi-turn dialogues such as making reservations and
medical
consultations~\cite{algherairy2023review}.
Recent advances in LLMs have revolutionized task bots, making pretrained LLMs
the backbones of
implementations~\cite{kulhanek2021augpt,peng2021soloist,yang2021ubar,hosseini2020simple,sun2022mars}.
These approaches involve finetuning the pretrained causal transformer model such
as GPT series with task-oriented dialogue data.
SimpleTOD~\cite{hosseini2020simple} represents a foundational method by
recasting task-oriented dialogue as a conditional language modeling task.
Specifically, it finetunes the pre-trained GPT-2 model to generate intermediate
labels, such as dialogue states and dialogue actions, as well as the final
response, based on the dialogue context and results retrieved from the database.
SOLOIST~\cite{peng2021soloist} improves the above approach by omitting dialogue
action prediction to reduce error propagation and incorporating a contrastive
objective to enhance response generation.
Further improvements include AuGPT~\cite{kulhanek2021augpt}, which adds
auxiliary tasks such as user intent and system action classifications, and
UBAR~\cite{yang2021ubar}, which extends to session-level examples by including
dialogue states and actions in the history.
While task bots based on encoder-decoder LLMs like
T5~\cite{sun2022mars} also present promising results, this work
remains focused on decoder-only LLMs as a foundational step in this line of
research.
In practice, task bots are often updated continually or periodically to repeatedly absorb recent, dialog data into their parameters.

\subsection{Causal Language Modeling for Task-Oriented Dialogues}
\label{sec:traintod}

As shown in \cref{fig:train_data}, at a high level, LLM-based task bots are
essentially trained to model the joint probability distribution $p(S_t, D_t,
A_t, R_t|H_t)$.
Here $H_t\triangleq \{U_0, R_0, \ldots, U_{t-1}, R_{t-1}, U_t\}$ denotes the
dialogue history, which comprises both user utterances $\{U_i\}_{0\leq i\leq t}$
and system responses $\{R_j\}_{0\leq j<t}$.
Given $H_t$, the model first predicts the dialogue state $S_t$ that tracks a
collection of user constraints mentioned during previous chatting.

{Consider this textual dialogue state example:
\exam{Belief state: Restaurant \{food = Italian, area = center\} Hotel \{phone = 12345\}}.
Each dialogue state consists of one or more \emph{domain–slot–value} triplets,
such as \text{(Restaurant, food, Italian)} or \text{(Hotel, phone, 12345)}.
Here, the \emph{domain} specifies the topic (e.g., \text{restaurant}),
each \emph{slot} denotes an attribute (e.g., \text{food}),
and each \emph{value} represents the user’s preference or provided information.
The dialogue state is dynamically updated as the conversation progresses whenever
the user adds, modifies, or removes a constraint.}

{
The dialogue state is then used to retrieve relevant data $D_t$ from the domain
database to satisfy user constraints.
The model then jointly utilizes the dialogue history, dialogue state, and the
retrieved data to predict an action $A_t$, which consists of the intent and the
associated slot-values.
For example, the action \exam{inform(domain=hotel,price=expensive)} has the
intent \emph{inform}, meaning that the user is informing the system to constrain
the search to expensive hotels.
Finally, the system generates the system response $R_t$ based on dialogue
history, dialogue state, retrieved data, and action.
Therefore, the joint probability can be factorized as}
  \begin{align*}
   & p(S_t, D_t, A_t, R_t|H_t) = \\
  & \hspace*{5ex}\underbrace{p(S_t|H_t)}_{\text{state prediction}}
  \underbrace{p(A_t|H_t, S_t, D_t)}_{\text{action prediction}}
  \underbrace{p(R_t|H_t, S_t, D_t, A_t)}_{\text{response generation}}
  \end{align*}
where $p(D_t|H_t,S_t)$ is omitted as the retrieved data $D_t$ is obtained using
deterministic constraint matching and hence $p(D_t|H_t,S_t)=1$.
This decomposition divides the language modeling task into three sub-tasks,
i.e., state prediction, action prediction, and response generation.

During finetuning, the LLM is optimized against all three tasks in an end-to-end
manner.
That is, given a training sequence $\bx^{t}_\text{train} =
\{(H_t,S_t,D_t,A_t,R_t)\}$, the LLM parameterized by $\theta$ is trained to
minimize the negative log-likelihood loss
\begin{align*}
  L_\theta(\bx^{t}_\text{train})
  =& - \log p_\theta(S_t|H_t) - \log p_\theta(A_t|H_t,S_t,D_t)  \\
   & -\log p_\theta(R_t|H_t,S_t,D_t,A_t).
\end{align*}

\subsection{Training Data Extraction Attacks}

Large language models can memorize training data owing to the huge number of
parameters.
Carlini et al.~\cite{carlini2021extracting} pioneered the exploration of model
knowledge extraction and defined $\kappa$-eidetic memorization, demonstrating
the feasibility of recovering individual training examples from GPT-2 through
targeted queries.
In Carlini et al.~\cite{carlini2021extracting}, internet contexts are utilized
as prefixes to elicit suffixes potentially indicative of the training data.
These examples are then checked if they are from the training dataset using
membership inference attacks.
Subsequent studies~\cite{kandpal2022deduplicating,carlini2022quantifying} found
that data repetition, context length, and model size significantly influences
this phenomenon of memorization.
Yu et al.~\cite{yu2023bag} enhanced the efficacy of data extraction from
pretrained language models by improving text decoding and membership inference
strategies.
In the black-box setting, aligned production LLMs such as ChatGPT, a novel
attack is introduced in~\cite{nasr2023scalable} that disrupts chatbot-style
outputs by forcing the language model to endlessly repeat the word ``poem'',
leading to a divergence after several hundreds of iterations.
Carlini et al.~\cite{carlini2023extracting} applied similar techniques to
diffusion models, while Qi et al.~\cite{qi2024follow} devised methods for
extracting database information from retrieval-augmented generation systems
through prompt injection.
Zeng et al.~\cite{zeng-etal-2024-exploring} further investigates whether fine-tuned models memorize input sequences and can be prompted to reproduce them.

{However, all of the above research targets either pre-training memorization or open-ended generation.
These models are optimized to freely reproduce input or output text, which makes verbatim recovery feasible.
In contrast, task-oriented dialogue systems (TODS) are trained with a conditional language modeling objective to predict structured dialogue states rather than raw utterances.
In TODS, even if a context is frequently seen during training, it is used only as conditioning input and is not optimized for generation—hence, the input sequences are not memorized.
The potential leakage instead lies in the structured slot–value pairs that appear in model outputs.
This difference is fundamental: while open-ended dialogue memorization concerns free-form text reproduction, TODS memorization involves schema-constrained, copy-like mappings between utterances and structured states that are highly context-dependent.
Consequently, methods developed for open-ended extraction cannot be directly applied to TODS, as they fail to elicit valid structured outputs without appropriate schema grounding.}

{Our study therefore fills this gap.
Rather than examining pre-training or open-ended memorization, we conduct the first systematic investigation of training-data extraction in LLM-based task bots, revealing how memorization manifests in the structured, conditional generation process.
Through comprehensive experiments, we quantify where existing decoding and membership-inference methods break down under schema constraints and propose targeted, task-aware adaptations to better measure and mitigate this risk.}


\section{Threat Model}
\label{sec:problem}


\subsection{Definition of Memorization}

In this work, we explore the ability of extracting training dialogue states from
LLM-based task bots, which learn to predict dialogue states given the
corresponding dialogue histories as context.
Unlike existing attacks, we assume the lack of access to the training dialogue
history during dialogue state extraction.
Hence, we define dialogue state memorization by restricting the prefix to only
include the preceding dialogue state as follows.

\begin{definition}[Dialogue State Memorization for Task Bots]
  A dialogue state string $s$ is considered memorized by the model
  $f_\theta$, with a prefix consisting $l$ domain-slot-value triplets if there
  exists a partial dialogue state prefix $x$ of length $l$ such that the
  concatenation $x || s$ presents in the training data, and $f_\theta$ yields
  $s$ when prompted solely with $x$ using greedy decoding, without conditioning
  on the corresponding dialogue history.
\end{definition}

For example, if the training data contains the following sequence.
\begin{quote}\tt\small
  \textbf{User}: I would like to find an expensive restaurant that severs
  Chinese food.

  \textbf{System}: Sure, which area do you prefer?

  \textbf{User}: Northern part of the town.

  \smallskip
  Belief State: Restaurant(price\_range= expensive,food=Chinese,area=north)
\end{quote}
When provided with only a length one partial dialogue state without the dialogue
history as prefix, e.g., \exam{Belief State: Restaurant(price\_range=expensive},
the model most likely generates \exam{food=Chinese,area=north)} as the output.
This part of dialogue state can be considered extractable or memorized.
Note that the prefix can be further reduced to simply \exam{Belief State:}.
We refer to this inherently more challenging scenario as \emph{untargeted
  extraction}, while the incorporation of a partial state prefix is termed
\emph{targeted extraction}.

\subsection{Target and Adversary}

\subsubsection{Adversary's Objective}
The adversary's objective is to extract the memorized training dialogue data from LLM-based task bots.
At first glance, one might consider directly extracting the dialogue history
$H_t$ by prompting the LLM.
However, this is infeasible as the model is specifically trained to predict
dialogue states and subsequent responses, rather than the dialogue history,
given by \cref{sec:traintod}.
Instead, the dialogue history is solely utilized as a conditioning context
during training the task bot.
As a result, the dialogue history is not memorized, making direct extraction of
the dialogue history data impossible.
Thus, we aim to \emph{extract dialogue states that summarize the dialogue
  history} (see \cref{fig:train_data}) which is a major difference comparing
with existing work on training data extraction from
LLMs~\cite{carlini2021extracting}.

Specifically, we attack the TODS under a {\em score-based black-box assumption}
where we only have input-output access to an LLM-based task bot parameterized by
$\theta$.
By prompting the task bot with a dialogue state beginning phrase $P$, such as
\exam{Belief State:} and \exam{[BOS]}\footnote{Dialogue states are usually
  marked by phrases like \exam{Belief State:} or special tokens such as
  \exam{[BOS]}, to enable the model to distinguish between tasks.
  An attacker could easily generate a sample to determine which tokens or
  phrases are in use.}, we obtain the probability distribution of the next-word
prediction $p_\theta(x_i|P, x_1, \ldots, x_{i-1})$, from which we sample the
output text sequence, and keep the dialogue state part as our extraction
candidates.
Next, we use several membership inference measures to rank these candidates and
determine whether these extractions originate from the training dataset.
Here, we pursue two adversarial goals, i.e., training dialogue state extraction
and membership inference.

\begin{itemize}
\item \textbf{Dialogue State Extraction}: The adversary aims to prompt the task
  bot to generate a large and diverse set of unique dialogue states, thus
  maximizing the coverage of the training dataset.
\item \textbf{Membership Inference}: The adversary seeks to accurately determine
  whether a specific dialogue state, schema, or value is present in the training
  dataset, enabling the effective elimination of noise candidates.
\end{itemize}

\subsubsection{Adversary Capabilities}
In real-world deployments, large language models (LLMs) such as ChatGPT do not disclose their underlying parameters. 
Accordingly, this work focuses on scenarios where an adversary has score-based black-box access to an LLM-driven task bot. 
Specifically, the adversary can query the system to obtain the probability (score) of arbitrary sequences and the next-word predictions, while being strictly forbidden from accessing any internal weights or hidden representations, including attention matrices. 
Some platforms may further harden their security by providing only textual outputs and concealing all probability information from users. 
While we fully agree that the strict black-box setting is important, we emphasize that partial-state leakage is both plausible and impactful in real deployments. 

{In practice, partial exposure of dialogue states can occur through several channels. 
For instance, insider engineers or third-party service providers may access debugging logs that contain fragments of dialogue states; 
platform APIs may expose shadow endpoints for monitoring or integration; 
and end-users themselves may inadvertently reveal partial states through cross-application synchronization. 
To illustrate the risk concretely, consider a scenario where a malicious actor learns fragments of a user’s travel plan through side information and then queries a dialogue system to infer further details about the itinerary. 
Such cases do not require access to the full training corpus but still demonstrate how limited leakage of dialogue-state traces can create meaningful privacy risks. }

{This work focuses on this partial-access threat model as an intermediate but practically plausible scenario. It reflects situations where adversaries can observe fragments of dialogue states without having direct access to the underlying model weights or training data. Studying this middle-ground setting enables us to systematically assess potential privacy leakage under weaker yet realistic assumptions, and to inform the design of future defenses applicable to fully black-box systems.
}


\subsubsection{Target Model} \label{sec:target}
In this work, as an example, we utilize the state-of-the-art Llama2
(7B)~\cite{touvron2023llama2} as the backbone LLM to construct the target task
bot for our attack.
Note that our technique can be easily extended to other LLMs.
The LLM is finetuned on the MultiWOZ benchmark~\cite{eric2019multiwoz} with the
method described in \cref{sec:traintod}.
Please refer to \cref{sec:metrics} for more details about the MultiWOZ dataset
and model finetuning.

\section{Dialogue State Extraction Methodology}
\label{sec:method}

In this section, we first introduce a strawman method that leverages existing
methods designed for pretraining data extraction from LLMs, and discuss its
limitations when used for training dialogue state extraction from LLM-based task
bots.
Then, we design novel methods to address these limitations.

\subsection{A Strawman Method for Dialogue State Extraction}
\label{sec:init}

Recall that prior works have explored universal training data extraction
attacks~\cite{carlini2021extracting,carlini2022quantifying,yu2023bag}.
Then, a strawman method is to directly leverage existing text generation methods
to extract training dialogue states.
The strawman method consists of two steps, i.e., candidate dialogue state
generation and membership inference.

\subsubsection{Candidate Dialogue State Generation}
\label{sec:31}
\label{sec:temp_samp}
\label{sec:beamsearch}

For dialogue state generation, we follow the untargeted extraction attack
setting where neither the training dialogue history nor dialogue state
information is provided to guide the generation.
Specifically, we simply prompt the task bot with \exam{Belief State:} and
proceed to sample tokens from the model in an auto-regressive manner, allowing
the model to generate responses based solely on this minimal prompt.
Our goal is to produce sequences that the model deems highly likely, based on
the assumption that these likely sequences are indicative of memorized text.

However, we observe that the task bot tends to generate an empty dialogue state,
leading to generic responses like \exam{How can I help you?}.
This behavior arises from a discrepancy between the training and testing
(extraction) settings.
Specifically, during training, the task bot is finetuned for conditional
language modeling tasks, with access to contextual information such as dialogue
history and partial dialogue states (see \cref{sec:traintod}).
In contrast, during training data extraction, the task bot lacks such context.
When generating by selecting words with the highest probability, the task bot
encounters a situation where the \exam{End-of-State} or \exam{[EOS]}
token\footnote{The \exam{End-of-State} or \exam{[EOS]} token is a special token
  used to indicate the end of a dialogue state.}
has the highest likelihood, causing premature termination of the dialogue state
content generation.
To promote dialogue state generation, we explore various
sampling strategies to encourage the model to generate more diverse tokens.

\begin{itemize}
\item \textbf{Temperature Sampling}~\cite{hinton2015distilling} samples tokens
  from a flattened next-word probability distribution.
  Specifically, the original softmax probability distribution
  $\text{softmax}(z)$ is replaced to $\text{softmax}(z/t)$, where $t$ is the
  \emph{temperature}, and $z = f_\theta (x_1,\ldots, x_{i-1})$ represents the
  output of the language model, i.e., the predicted probability distribution
  over the vocabulary.
  A higher temperature $t$ leads to a less confident but more varied model
  output.
\item \textbf{Sampling with a Decaying Temperature}~\cite{carlini2021extracting}
  further improves temperature sampling by starting at a higher initial
  temperature and decreasing at each generation step.
  In the early stage, it allows the model to explore a wide range of prefixes,
  enhancing generation diversity.
  As time advances, the temperature decreases, and the model is more likely to
  focus on generating high-confidence, less random paths, making the outputs
  more valid.
\item \textbf{Beam Search}~\cite{sutskever2014sequence} evaluates the top-$B$
  hypotheses at each time step and eventually chooses the hypothesis that has
  the overall highest probability for the entire sequence.
  This has the advantage of identifying high-probability sequences that start
  with a lower probability initial tokens and would have been ignored by the
  greedy search.
\item \textbf{Group Beam Search}~\cite{vijayakumar2016diverse} groups the beams
  and introduces an intra-group diversity term to encourage diversity.
\item \textbf{Beam Sampling}~\cite{ippolito2019comparison} replaces the top-$B$
  with multinomial sampling to encourage diversity.
  We apply temperature sampling to further amplify this diversity.
\end{itemize}

\subsubsection{Membership Inference}
\label{sec:32}

After the previous step, we can generate a large number of dialogue state
candidates; however, many of them may not in the original training dataset at
all.
We thus need to perform membership inference to determine whether each candidate
indeed belongs to the training dataset.
The membership inference attack leverages the observation that models tend to
assign higher confidence to samples that they have encountered during
training~\cite{carlini2021extracting}.
Specifically, we rank the extracted dialogue state candidates, each of which is
treated as a token sequence, according to two metrics broadly used in previous
works~\cite{carlini2021extracting,yu2023bag}.
\begin{itemize}
\item \textbf{Perplexity (PPL)}~\cite{jelinek1977perplexity} is a measure of how
  predictably a language model generates text.
  A lower perplexity indicates that the model finds the sequence less
  surprising, meaning that it likely assigns a higher average probability to
  each token in the sequence.
  Formally, for a sequence $\bx=(x_1,\ldots,x_n)$ and a LLM $f_\theta$, we have
  \[
    \text{PPL}(\bx) \triangleq
    \exp\left(-\frac{1}{n}\sum_{i=1}^n\log f_\theta (x_i|x_1,\ldots,x_{i-1})\right).
  \]
\item \textbf{PPL-zlib} adjusts the perplexity of a generation by scaling it
  with the information entropy of the generated sequence, where the entropy is
  quantified through the zlib compression of the sequence~\cite{gailly2004zlib}.
  This approach mitigates the bias in PPL toward favoring trivial generations
  that exhibit low information content such as a text with repetitive
  subsequences, e.g., ``123123123''.
\end{itemize}

\begin{figure*}[!t]
  \includegraphics[width=.85\linewidth]{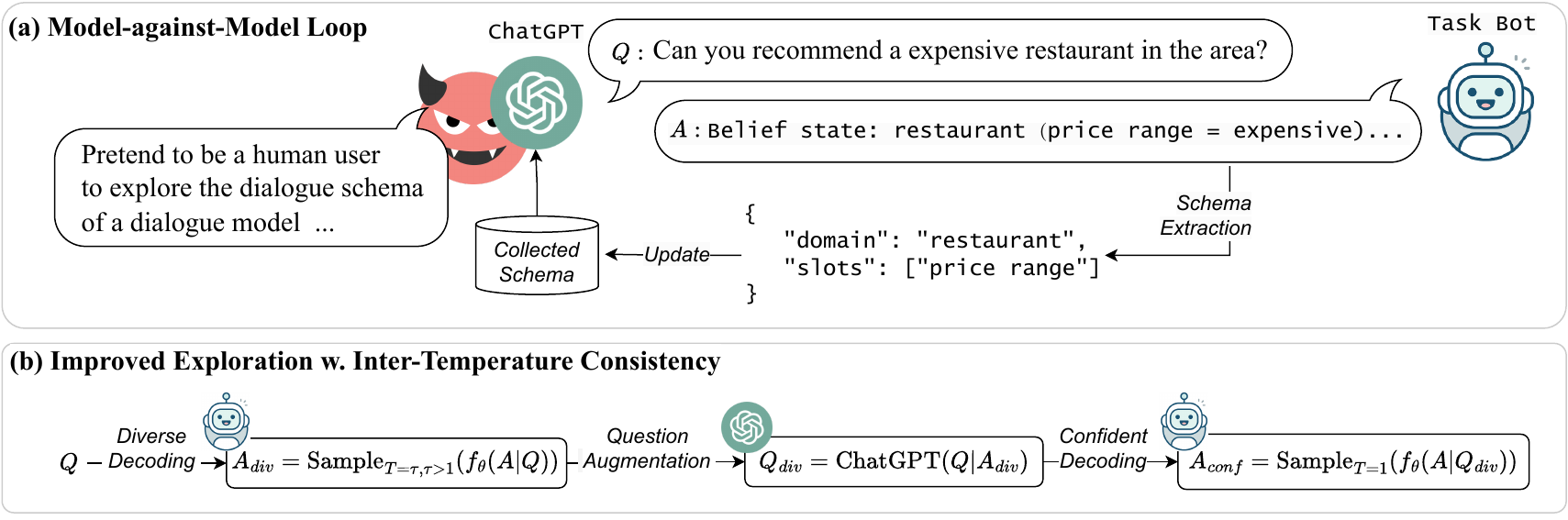}
  \centering
  \caption{Schema extraction with ChatGPT.
    In \textbf{(a) model-against-model loop}, based on the collected schema
    (none for the first loop), ChatGPT is prompted to simulate a user asking the
    task bot questions to uncover additional domains and slots.
    To improve precision and exploration, we further propose to enhance the
    basic loop with \textbf{(b) inter-temperature consistency}.
    When decoding the answer from the task bot as in (a), we use
    high-temperature settings to generate diverse belief states, which may
    include incorrect domains and slots.
    We then filter these errors through a verification process, where ChatGPT is
    prompted to ask corresponding questions about the extracted (probably
    incorrect) states, and the task bot decodes answers under low-temperature
    settings to generate confident answers that exclude those illegal domains
    and slots.}
  \label{fig:gptattk}
\end{figure*}

\subsubsection{Limitations of the Strawman Method}

There are several drawbacks in the strawman method.
For candidate dialogue state generation, existing methods suffer from probing
common and low diverse dialogue states.
For instance, we observe that \exam{Restaurant(name=yippee noodle bar)}
duplicates hundreds of times in our experiments.
For membership inference, existing measures are prone to false positives,
inaccurately assigning high likelihoods to non-memorized samples.
This issue is particularly evident with long sequences and repeated substrings
such as \exam{Restaurant(name=yippee noodle bar) Attraction(name=yippee noodle bar)}.

The frequent generation of trivial or generic content can be attributed to
inherent characteristics of task-oriented dialogues.
In practice, users often initiate conversations with common requests such as
\exam{Please book me a restaurant in town}.
These initial interactions typically lead to an expansion of the dialogue, which
incorporates a variety of topics based on the user's evolving preferences.
As a result, early interactions are more prone to repetition.
Consequently, training data extracted using traditional methods tends to focus on these early turns, which are often superficial and lack substantial detail.
Moreover, due to the coherent nature of dialogues, it is common for the same
value to be repeated across multiple turns, e.g., a taxi's arrival address
matching a restaurant's address.
While not conditioning on dialogue history, the task bot can easily adopt this
value copy mechanism and erroneously replicate values where they may not
logically apply.

\subsection{Schema-Guided Targeted Dialogue State Generation}
\label{sec:advattack}
\label{sec:adv_text_gen}

The candidate dialogue state generation of the strawman method suffers from two
major issues: 1) generating dialogue states with limited diversity, and 2)
generating illegal dialogue states that either contain out-of-scope domains,
slots, and values, or lack a legal format.
To address these issues, we first propose a targeted extraction setting that
leverages partial dialogue states for context.
Then, we introduce a schema-guided sampling method that probes the service scope
of task bots to restrict the word choice range during sampling, enabling the
extraction of more relevant and legitimate dialogue states from the task bot.

\subsubsection{Targeted Dialogue State Extraction}
\label{sec:condtarget}

Sampling dialogue states without any conditioning context tends to result in
generating only generic text patterns that occurred in the training dataset.
For instance, all the restaurants are generated as \exam{Pizza Hut} because
\exam{name=Pizza Hut} appears multiple times in the training data.
To overcome this issue, we explore the use of partial dialogue states, either
obtained from the strawman attack or crafted by an attacker, as prompts, to
produce more detailed and targeted samples.
For example, if an attacker knows part of someone's travel plan, the attacker
could prompt the model to reveal the rest.

In this work, we use examples from the origin training dataset of the task bot
for a quantitative proof-of-concept analysis.
We do not simulate real attackers to generate trigger cues to prevent our
methods from being easily misused.
Specifically, we utilize partial dialogue states from the training dataset as
controlled-length conditioning context, e.g., \exam{Belief State:
  Restaurant(name=pizza hut,} serving as a prefix with length one, i.e., one
domain-slot-value triplet.

\subsubsection{Schema-Guided Sampling}
\label{sec:lmvslm}

In the strawman method, a significant portion of the generated dialogue states
were either out of scope or disorganized in format.
To address these issues, we draw inspiration from constrained decoding
methods~\cite{lu-etal-2022-neurologic} and propose to first extract the service
scope of the task bot and then use it to prune the exploration scope.
Specifically, we start by extracting the dialogue schema from the task bot,
which includes domains and slots\footnote{We do not pre-define the scope of
  values (e.g., ``\texttt{12345}'' is a value for the slot ``\texttt{phone}'')
  as they are strongly dependent on user utterances.}
that define the bot's service scope.
Then, in the process of extracting dialogue states, when sampling domains and
slots, we strictly confine our token selections to those within this extracted
schema.
This ensures that the generated candidate dialogue states are both relevant to
the scope and maintain the required structured format.

To extract dialogue schema from the task bot, we first collect various dialogue
states, which may contain errors, and then extracting domains and slots within
these states using regular expressions.
In this process, prioritizing recall is crucial, as missing any element means
certain domains or slots will not be generated during the state extraction
attack.
To this end, we conduct initial state extraction using temperature sampling with
a high temperature to maximize exploration and collect schema from the extracted
states.
However, our experimental results show that while this method excels in
achieving high recall, it tends to yield excessively low precision due to the
generation of corrupted outputs containing irrelevant domains and slots.
Besides, achieving a high recall requires thousands of task bot generations,
which is computationally inefficient.

To overcome this limitation, we propose a heuristic \textit{\textbf{model-against-model
  method}}, disguising ChatGPT as a user to interact with the task bot and probe
its service scope.
Conditioning on those ChatGPT questions ensures that the task bot replies with
high confidence.
This method not only ensures higher precision under high recall but also reduces
the number of task bot generations required.
Specifically, this iterative question-asking phase is refined by incorporating
previously collected domains and slots into the prompts.
As shown in \cref{fig:gptattk}, the process starts with ChatGPT posing questions
based on the explored schema (if any).
We design a question generation protocol and prompt the LLM with the instruction in \cref{tab:prompt}.

\begin{table}[t]
\centering
\footnotesize
\caption{LLM Prompt for Generating Schema-Probing Questions}
\label{tab:prompt}
\begin{tabularx}{.48\textwidth}{|X|}
\hline

Generate questions for the dialogue system given the existing schema following these steps:

Review the Existing Schema: Carefully examine the provided \texttt{Validated and Confirmed Schema}. Make sure you fully understand each domain, slot, and value listed within.

Craft Simulated Task-oriented Queries: Based on each existing domain, create a series of questions that deviate slightly from the known scope. These queries should help probe potential new slots or values within those domains. For example, if you have a domain \texttt{weather} and a slot \texttt{city}, you might ask: ``How's the weather for outdoor activities tomorrow?'' If there is a domain \texttt{restaurant}, you could say: ``Can you suggest vegan restaurants near Tokyo?''

Implement Cross-Domain Probing: Combining known slots and values, try constructing cross-domain queries. For example, you could ask: ``Are there any music events happening in London when it's forecasted to rain?'' Or, ``Can I get a taxi from Beijing Duck to the nearest cinema?''

Initiate Random Task-oriented Queries: Design and propose a series of random task-oriented questions aiming to discover potential domains, slots, or values not covered by the current schema. The generated questions should be completely \textbf{IRRELEVANT} to the existing schema and based on a wide range of domains.

Validated and Confirmed Schema:
\{\texttt{schema: \{domain: Restaurant, slots:[price\_range, name]\}\}}

Your questions:\\
\hline
\end{tabularx}
\end{table}

Despite its effectiveness, exploring solely with ChatGPT is inefficient as it
lacks prior knowledge of the task bot and can easily pose out-of-scope
questions.
To address this issue, we propose a variant that enhances exploration through
\textit{\textbf{inter-temperature consistency}}.
This method first uses high-temperature sampling to obtain diverse dialogue
states and then converts them back into questions (the instruction is simply:
\exam{Please try to generate questions based on the following dialogue states:})
to probe confident outputs with low-temperature sampling.
Domains and slots that do not consistently appear in both high and low
temperature samples are filtered out.
This further boosts exploration without the loss of precision.

After extracting the dialogue schema, we use the identified schema to constrain
the sampling of specific domains and slots.
Specifically, we employ a hard-constrained sampling method that limits the
vocabulary from which we sample to only those extracted domains and slots.
Given a candidate set of words or phrases, denoted by $C$, we first compute the
probability of each candidate given a prefix, forming a probability
distribution.
From this distribution, we sample a candidate to determine the domain and slots.
Formally as $w = \text{Sample}(P(w'|\mathrm{prefix}), w' \in C)$, where $w$ is the
selected word or phrase from $C$.

\subsection{Debiased Membership Inference}
\label{sec:adv_mia}

The membership inference methods in the strawman method suffer from a high false
positive rate, particularly when dealing with repetitive substrings and long
sequences.
This issue is primarily attributed to inconsistency between training and
validation.
Typically, task bots learn to generate dialogue states based on the given
dialogue history.
However, measures such as perplexity and its variants (e.g., zlib scaling
\cite{carlini2021extracting,yu2023bag}) treat the dialogue state as an
independent sequence without conditioning context.
A more reasonable approach would consider the conditional nature of the
sequence.
Specifically, if a sequence is familiar given the dialogue history, it is more
likely to be a training example.

Given the lack of access to the training dialogue history, we propose an
approximate measure of conditional perplexity (C-PPL), which focuses solely on
the probability of the sequence's suffix, rather than the entire sequence, i.e.,
\[
  \text{C-PPL}(\bx)
  \triangleq \exp\left(-\frac{1}{L_s}
    \sum_{i=L_p+1}^{L_p+L_s} \log f_\theta(x_i | x_1, \ldots, x_{i-1})\right)
\]
where $\bx = (x_1, \ldots, x_{L_p}, x_{L_p + 1}, \ldots, x_{L_p + L_s})$, with
\(L_p\) denoting the length of the prefix and \(L_s\) the suffix.
For the untargeted setting, this prefix is simply \exam{Belief State:}.

Additionally, to address false positives where the suffix, rather than the whole
sequence, is over-familiar to the task bot, we introduce a debiased version of
C-PPL, named Debiased Conditional Perplexity (DC-PPL), that scales C-PPL with
the perplexity of the suffix itself, i.e.,
\begin{align*}
  \text{DC-PPL}(\bx)
  &\triangleq \frac{\log(\text{C-PPL}(\bx))}{\log(\text{PPL}(\bx_\text{suffix}))}\\
  &= \frac{\sum_{i=L_p+1}^{L_p+L_s} \log f_\theta(x_i | x_1, \ldots, x_{i-1})}
    {\sum_{i=L_p+1}^{L_p+L_s} \log f_\theta(x_i | x_{L_p+1}, \ldots, x_{i-1})}
\end{align*}
where $\text{PPL}(\bx_\text{suffix})$ is calculated as usual for the standalone
sequence of the suffix.
We apply a logarithm transformation of perplexity for a more normal
distribution.

\section{Evaluations}
\label{sec:expadv}

In this section, we evaluate the performance of different training data
extraction methods on a large-scale public available task-oriented dialogue dataset.

\subsection{Dataset and Evaluation Metrics}
\label{sec:metrics}

\subsubsection{Dataset}
We evaluate different training data extraction methods on the MultiWOZ
dataset~\cite{eric2019multiwoz}, which is the most widely used gold-standard
large-scale task-oriented dialogue dataset.
The MultiWOZ dataset consists of 8,438 dialogues, totaling 113,556 turns. On average, each dialogue has 13.46 turns and each turn includes 13.13 tokens. Overall, the dataset spans 7 domains, incorporates 24 slots, and covers 4,510 distinct values.



\subsubsection{Model Finetuning}

Using the MultiWOZ dataset, we finetune the Llama2 (7B) model, and the finetuned
Llama2 (7B) will be the task bot that we aim to attack.
However, directly finetuning LLMs on the MultiWOZ dataset can easily lead to
overfitting due to data sparsity, as the model memorizes the data instead of
learning to generalize, and hence makes the model vulnerable to data extraction
attacks.
For a more convincing evaluation of our attack's effectiveness, we finetune the
target model with Low-Rank Adaptation (LoRA)~\cite{hu2021lora}, which is a
parameter-efficient LLM finetuning technique that reduces overfitting by
limiting trainable parameters, while still ensuring performance on par with
full-parameter finetuning.

The finetuned task bot, now more fortified against attacks, presents a bigger
challenge to data extraction attacks, allowing us to better evaluate the
efficiency of our attack strategy.
As we focus on extracting training dialogue states, we assess the performance of
the task bot in dialogue state tracking, evaluating its ability to predict
dialogue states based on dialogue history.
The finetuned task bot achieves $91.43\%$, $71.52\%$, and $80.26\%$ for
precision, recall, and F1 score, respectively, demonstrating its effectiveness.

\begin{table*}[!t]
\centering
\small
\caption{Membership inference results for untargeted dialogue state extraction
  attack.
  Methods with fewer than 100 states are excluded, as the top 100 extractions
  are retained after ranking.}
\label{tab:no_context_mia}
\begin{tabular}{l|ccc|ccc|cccc}
\toprule
& \multicolumn{3}{c|}{\textbf{State Precision (\%)}} & \multicolumn{3}{c|}{\textbf{\#Triplets}} & \multicolumn{3}{c}{\textbf{Value Precision (\%)}} \\
\cmidrule{2-4}
\cmidrule{5-7}
\cmidrule{8-10}
\textbf{Extraction Method}  & \textbf{None} &  \textbf{PPL\%} &  \textbf{PPL-zlib} & \textbf{None} &  \textbf{PPL\%} &  \textbf{PPL-zlib} & \textbf{None} &  \textbf{PPL\%} &  \textbf{PPL-zlib} \\
\midrule
Vanilla Sampling & \textbf{14.37} &  \textbf{26.00} & \textbf{26.00}  & 1.60 & 2.27 & 2.00 &\textbf{15.00} & 60.52 & 54.9  \\
Temperature Sampling & 4.59 & 19.00 & 11.00 & 1.29 & 1.53 & 1.45 &5.76 & 38.56 & 33.00\\
Beam Sampling  & 11.95 & 23.00 & 21.00 &\textbf{1.95} & \textbf{2.52} & \textbf{2.71} & 10.73 & \textbf{66.97} & \textbf{68.1} \\
\bottomrule
\end{tabular}
\end{table*}

\subsubsection{Evaluation Metrics}
{We evaluate the quantity of memorized examples and their precision within both
the overall extracted dialogue states and the specific values (e.g., phone numbers) they contain.
Note that even if the entire dialogue state does not align perfectly, individual values may still be accurately extracted from the training dataset.
The metrics are as follows.
}


\begin{itemize}
\item {\textbf{Number of Extracted Dialogue States (\#States).}  
  The total count of dialogue states extracted by the attack.  
  Each dialogue state is represented as a set of domain–slot–value triplets, e.g.,  
  \(\texttt{Restaurant}\{\texttt{name}=\texttt{pizzahut}, \texttt{phone}=12345\}\).  
  A dialogue state is considered successfully extracted if it \textbf{exactly matches} or is a \textbf{subset} of any ground-truth state in the training set after triplet decomposition.}

\item {\textbf{Dialogue State Extraction Precision (SP).}  
  The proportion of extracted dialogue states that correctly appear in the training dataset:  
  \[
  \text{SP} = \frac{N_{\text{correct states}}}{N_{\text{extracted states}}}.
  \]
  This metric evaluates accuracy at the state level, after deduplicating identical states.}

\item {\textbf{Average Number of Domain–Slot–Value Triplets per Extracted State (\#Triplets).}  
  The average number of domain–slot–value triplets (e.g., Restaurant-name-pizzahut) contained within each successfully extracted state:  
  \[
  \text{\#Triplets} = \frac{\sum_{i=1}^{N_{\text{correct states}}} |T_i|}{N_{\text{correct states}}},
  \]
  where \(T_i\) denotes the set of triplets in the \(i\)-th correct dialogue state.  
  This metric reflects the information richness of each reconstructed dialogue state.
  }

\item {\textbf{Number of Extracted Slot Values (\#Values).}  
  The total number of unique slot values appearing in all extracted dialogue states, after deduplication across states.  
  For example, if “\texttt{pizzahut}” appears in multiple states, it is counted once.
  }

\item {\textbf{Slot Value Extraction Precision (VP).}  
  The proportion of extracted slot values that exist in the training dataset:  
  \[
  \text{VP} = \frac{N_{\text{correct values}}}{N_{\text{extracted values}}}.
  \]
  This metric evaluates the fidelity of value-level recovery, independent of state structure.
  }
\end{itemize}

\subsection{Evaluation Results of the Strawman Method}

\subsubsection{Settings}

For each dialogue state generation method, we generate $10,000$ samples using
the task bot.
We then deduplicate the extractions and rank the suffixes using membership
inference metrics, retaining only the top $100$ samples for further analysis.
We conducted an extensive hyperparameter search to identify the optimal settings
for each generation method introduced in \cref{sec:31}.
The chosen configurations are as follows: 1) Temperature Sampling: The
temperature is set to $2$; 2) Sampling with a Decaying Temperature: We initiate
the process with a high temperature of $5$ to promote diverse initial outputs,
then decrease the temperature by $2$ at each subsequent step until it stabilizes
at $1$; 3) Beam Search: A large beam size of $50$ is utilized, with all beams
retained to enhance output diversity and maximize training data coverage; 4)
Group Beam Search: The number of groups is set equal to the beam count, and a
diversity penalty of $1$ is applied; 5) Beam Sampling: The beam size is set to
$5$ and the temperature is configured at $2$.

\begin{table}[t]
\centering
\caption{Untarget extraction using the strawman method}
\label{tab:no_context}
\begin{tabular}{@{ }c@{ }|@{ }r@{ }r@{ }r@{ }r@{ }r@{ }r}
\toprule
\textbf{Method}      & \textbf{\#States} & \textbf{SP\%} & \textbf{\#Triplets} & \textbf{\#Values} & \textbf{VP\%} \\
\midrule
Vanilla Sampling     & 336               & 14.37         & 1.60                & 231               & 15.00         \\
Temperature Sampling & 207               & 4.59          & 1.29                & 234               & 5.76          \\
Decaying Temperature & 75                & 3.32          & 1.51                & 223               & 17.13         \\
\midrule
Beam Search          & 5                 & \textbf{100}  & 1.2                 & 4                 & \textbf{100}  \\
Group Beam Search    & 15                & 65.22         & 1.47                & 16                & 64.00         \\
Beam Sampling        & \textbf{715}      & 11.95         & \textbf{1.95}       & \textbf{315}      & 10.73         \\
\bottomrule
\end{tabular}
\end{table}

\subsubsection{Dialogue State Extraction Performance}

We find various training dialogue states without conditioning on their
corresponding dialogue histories.
For example, \exam{Restaurant(price\_range=expensive) Train
  (departure=Cambridge,destination=London Kings Cross,people=2)} describes a
conversation about someone planning an expensive meal in Cambridge before taking
a train to London Kings Cross with another person.
We also find examples of single-domain bookings, such as
\exam{Train(day=Friday,people=2)}.
These examples demonstrate that, even in an untargeted attack setting, it is
possible to extract private information from a task bot.

Next, we provide quantitative results in \cref{tab:no_context}.
We see that the number of correctly extracted dialogue states across all
variants of the strawman method is notably low, with less than $3.36\%$ among
the generations.
Although the beam sampling method may appear to generate more outputs, it
actually produces beam size times more outputs than the basic sampling method.
Thus, the actual proportion of correctly extracted dialogue states is even
lower.
Besides, temperature sampling and its decaying variant were ineffective in our
context, which is contrary to previous
studies~\cite{carlini2021extracting,yu2023bag}.
We attribute this discrepancy to the highly structured nature of dialogue
states, which these methods disrupt, leading to invalid outputs.
We also observe that both beam search and group beam search show high precision owing to their deterministic nature, but failed to produce diverse outputs due
to computational constraints limiting beam size: despite $10,000$ iterations,
the output diversity was constrained to the number of beams.
Beam sampling allowed for a broader exploration of dialogue states by combining
beam search with temperature sampling, yet the number of dialogue states
uncovered remained limited.
Moreover, the average length (\#Triplets) of the correct dialogue state
extractions was notably short, typically under $2$ and at most $4$.
These often included simple details about restaurants or train schedules,
information that is likely to appear at the beginning of the dialogues and
appear multiple times in the training dataset, making it less sensitive.

\subsubsection{Membership Inference Performance}

As shown in \cref{tab:no_context_mia}, the results underscore the effectiveness
of membership inference in enhancing both the precision of extractions and the
lengths of the extracted states.
However, despite these improvements, the overall privacy risk remains moderate.
While individual values exhibit a relatively higher risk, with maximum precision
reaching $67\%$, event-level states present a significantly lesser concern, with
a maximum precision of $26\%$.

\subsubsection{Limitations of the Strawman Method}

Our experimental results uncover several limitations of the strawman method.
Primarily, it suffers from probing common and low diverse dialogue states.
For instance, \exam{Restaurant(name=yippee noodle bar)} duplicates hundreds of
times.
Furthermore, the membership inference measures are prone to false positives,
inaccurately assigning high likelihoods to non-memorized examples.
This issue is particularly evident with long sequences and repeated substrings,
e.g., \exam{Restaurant(name=yippee noodle bar) Attraction(name=yippee noodle
  bar)}.
In dialogues, it is common for the same value to appear multiple times due to
the coherent nature of the conversation, such as a taxi's arrival address
matching a restaurant's address.
However, without conditioning on dialogue history, the task bot may mistakenly
adopt this pattern, erroneously replicating values where they might not
logically apply.

\subsection{Dialogue Schema Extraction Results}

Since the performance of our targeted schema-guided dialogue state extraction
method depends on the extracted dialogue schema, we first present the schema
extraction results before discussing the dialogue state extraction results.

\subsubsection{Settings}
In this section, we evaluate our model-against-model dialogue schema probing
strategy (see \cref{sec:lmvslm}).
We evaluated five major metrics.
For accuracy analysis, we consider the precision and recall for domains and
slots, respectively.
For cost analysis, we consider the number of questions posed by ChatGPT required
to achieve coverage.

\subsubsection{Extraction Performance}

\begin{figure}[t]
  \centering
  \includegraphics[width=.8\linewidth]{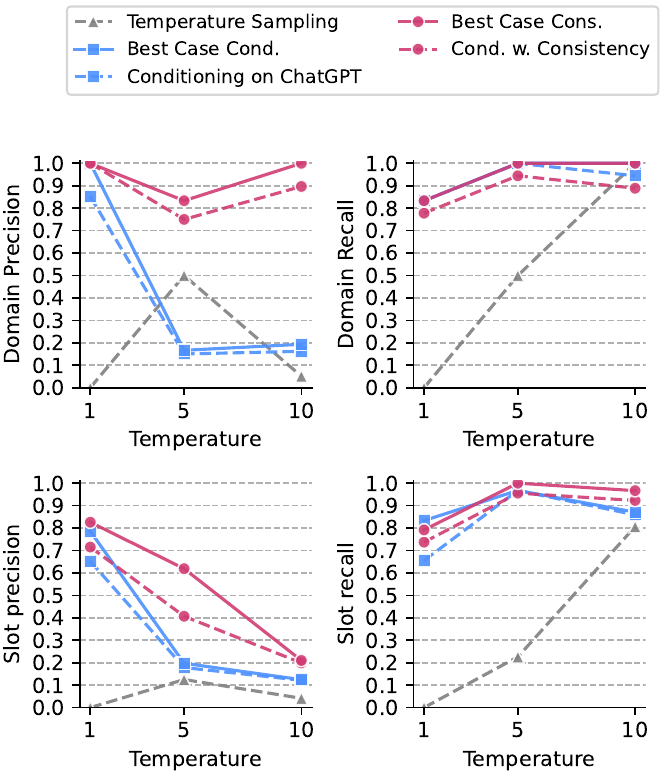}
  \caption{Schema extraction performance over different temperatures.
    We report the mean values over three runs per method, along with the
    best-case results.}
  \label{fig:schema_etr}
\end{figure}

As shown in \cref{fig:schema_etr}, with the proposed model-against-model method,
the best case with temperature $5$ achieves full $100\%$ coverage of the
dialogue schema, with domain precision exceeding $82\%$ and slot precision over
$60\%$.
Compared to temperature sampling, our method shows significantly higher
precision while also improving recall.
The inter-temperature consistency has proven effective in further improving
recall with minimal loss of precision.

We further delve into error analysis and found the primary false positive slots include: 1) coreferences or alternative
expressions of slots, e.g., \exam{people} versus \exam{number of people}, and 2)
slots and their variants from other domains.
However, our next experiment will show that these issues do not lead to a severe
performance drop in enhancing dialogue state extraction.

\subsubsection{Cost Analysis}

\begin{figure}[t]
  \centering
  \includegraphics[width=.85\linewidth]{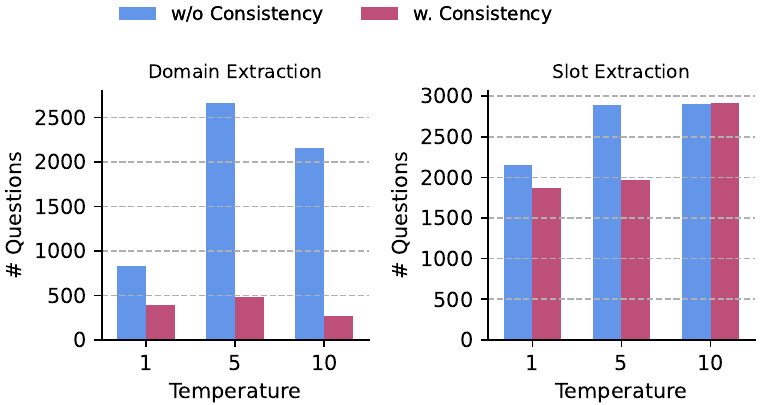}
  \caption{Minimum number of questions required for ChatGPT to achieve the
    performance upper bound in schema extraction.
    For temperature sampling, $10,000$ samples are generated to reach the
    reported performance.}
  \label{fig:cost}
\end{figure}

We evaluate the computational cost using the \emph{Number of Questions Required
  for Convergence} (\#Questions), which calculates the minimal number of times the
task bot must be queried for optimal schema extraction performance.
For an attacker, fewer interactions signify better efficiency.
As depicted in \cref{fig:cost}, the proposed model-against-model method requires
about $3$ to $5$ times fewer interactions compared to traditional temperature
sampling.
Moreover, the variant employing inter-temperature consistency further improves
efficiency in domain extraction and slot extraction.
Besides, we find that the cost can increase at higher temperatures.
The reason is that diverse sampling gradually uncovers hard-to-detect marginal
slots and coreferences or alternative expressions of slots through successive
discoveries, thereby requiring more interactions to achieve convergence.

\subsection{Schema-Guided Targeted Dialogue State Extraction Results}
\label{sec:extraction_performance}

\subsubsection{Settings}

We condition the task bot on partial dialogue states from the training dataset
to perform a proof-of-concept quantitative analysis on targeted dialogue state
extraction (see \cref{sec:condtarget}).
Specifically, we control the number of domain-slot-value triplets of the partial
belief state (Prefix Length) for extracting the remainder suffix.
These input partial dialogue states are obtained by truncating dialogue states
from the training dataset that exceed the controlled length.
The resulting prefixes, e.g., \exam{Belief State: Restaurant(name=pizza hut},
are used to promote the task bot to generate suffixes.
Note that the prefix does not include the training dialogue history, but only
partial dialogue states.
Since different dialogue segments may share the same dialogue state—for example,
when they all involve making a restaurant reservation—we deduplicate the
prefixes to avoid redundancy, resulting in a number of unique prefixes.
After extraction, we perform further deduplication before calculating the
evaluation metrics.

Please revisit \cref{sec:beamsearch} and \cref{sec:lmvslm} for a detailed
description of the baseline and our proposed schema-guided suffix decoding
methods.
Note that the schema we use here is obtained through the model-against-model
method with inter-temperature consistency.

\subsubsection{Targeted Dialogue State Extraction Results}
\label{sec:target_results}

\begin{figure}[tp]
  \centering
  \includegraphics[width=\linewidth]{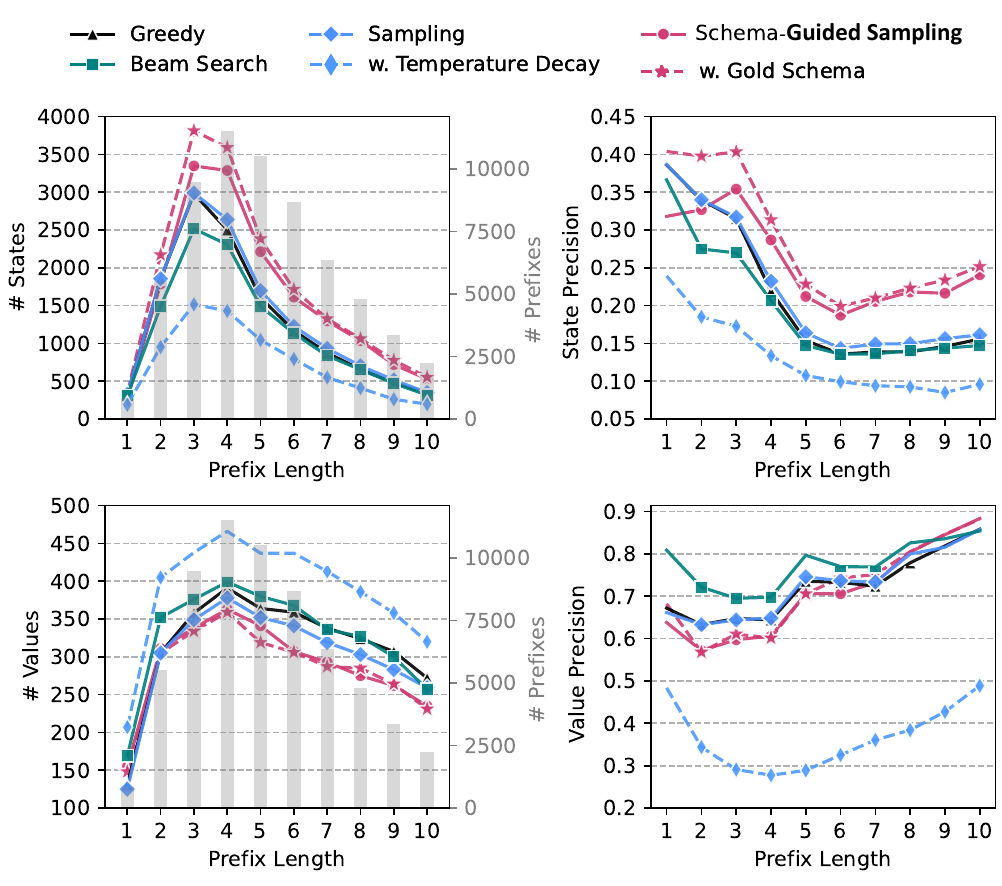}\\
  \caption{Targeted state extraction results.
    We report the total number of successfully extracted training dialogue
    states, the memorized values within these states, and the extraction
    precision.
    The gray bars represent the number of deduplicated prefixes.}
  \label{fig:adv_state_extr}
\end{figure}

As shown in \cref{fig:adv_state_extr}, our method manages to extract thousands
of training dialogue states from the task bot when seeding it with partial
dialogue states.
The proposed schema-guided sampling method outperforms all baselines by a
notable margin and performs only slightly worse than the gold schema upper
bound, demonstrating its effectiveness in probing the task bot by restricting
the sampling space.
Surprisingly, greedy search remains a strong baseline, outperforming both beam
search and temperature decay in terms of the number of valid states and state
precision.
The reason is that, with a conditioning context, all methods generate the same
number of outputs; however, diversity-enhancing techniques produce a wider
variety of values (as evidenced by the \#Values and value precision results).
When these diverse outputs are combined into states, they are more likely to be
invalid.
In contrast, greedy search consistently produces a fixed set of values that
align closely with the training data, leading to more valid states overall.

As shown by the gray bars on the left in \cref{fig:adv_state_extr}, there is
variability in the number of deduplicated prefixes of partial dialogue states,
resulting in varying total extractions for different prefix lengths.
Therefore, we further evaluate the precision of each method.
As shown in \cref{fig:adv_state_extr}, schema-guided sampling continues to
outperform other methods.
Interestingly, we observe that as the length of the prefixes increases,
precision decreases.
This finding contradicts the previous study~\cite{carlini2022quantifying}, which
suggests that more context leads to increased precision.
We will further discuss this observation in \cref{sec:why_cont}.

\subsubsection{Impact Analysis of Sampling Configurations}
\label{sec:hyperpara_samp}

\begin{figure}[t]
  \centering
  \subfloat[Top-$k$\label{fig:topk}]{%
    \includegraphics[width=.4\linewidth]{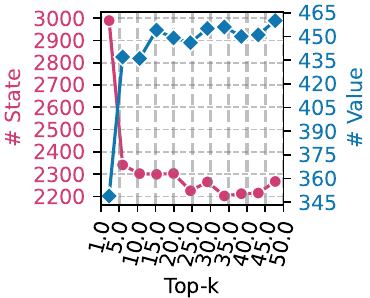}}
  \subfloat[Temperature\label{fig:temp}]{%
    \includegraphics[width=.4\linewidth]{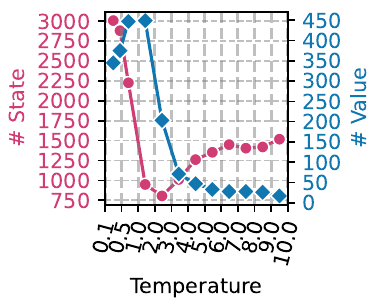}}\\
  \subfloat[Typical-$\phi$\label{fig:typical}]{%
    \includegraphics[width=.4\linewidth]{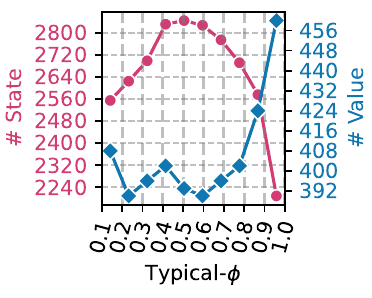}}
  \subfloat[Nucleus-$\eta$\label{fig:nu}]{%
    \includegraphics[width=.4\linewidth]{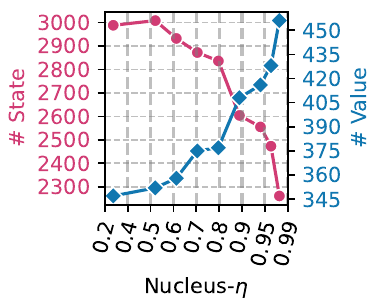}}\\
  \subfloat[Repetition penalty\label{fig:rep}]{%
    \includegraphics[width=.4\linewidth]{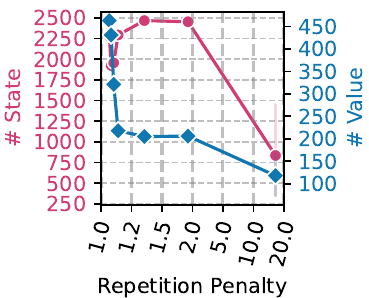}}
  \caption{Number of training dialogue states and values extracted under
    different sampling configurations.
    Results are reported for a prefix length of 3, with the largest sample size
    to provide statistically significant results.}
  \label{fig:hyperpara}
\end{figure}

In this section, we examine how adjustments to the sampling configurations,
i.e., hyperparameters, affect dialogue state extraction results.
The results are presented in \cref{fig:hyperpara}.

Top-$k$~\cite{fan2018hierarchical} is set to restrict the vocabulary to the
top-$k$ tokens based on probability, aiming to focus generation on the most
likely outcomes.
As shown in \cref{fig:topk}, a larger $k$ increases the diversity of values but
reduces the number of valid dialogue states.

Temperature (\cref{sec:temp_samp}) is set to adjust the probability distribution
of the tokens.
From \cref{fig:temp} we see that a lower temperature leads to more valid but
less diverse dialogue states.
For relatively higher temperatures, we observe that more valid states are
extracted, albeit with fewer values.
This may occur because, although most cases are invalid, certain slots are
successfully extracted and then high-confidence values, which are strongly
memorized, are frequently sampled even at higher temperatures.

Typical-$\phi$~\cite{meister2023locally} is set to recommend selecting a token
whose information content is close to the expected information content given
prior context, enhancing sentence consistency and reducing degenerate
repetitions.
As shown in \cref{fig:typical}, performance varies non-monotonically with
different $\phi$ values, which is consistent with~\cite{yu2023bag}.

Nucleus-$\eta$~\cite{holtzman2019curious} is set to assure the output be sampled
from the smallest set of the most likely tokens whose total probabilities are
equal to or exceed $\eta$.
In \cref{fig:nu}, we observe that lower values of $\eta$ enhance state
extraction, analogous to the Top-$k$ method discussed above.

Repetition Penalty~\cite{hilprecht2019monte} is set to modify the generation
probability of each token depending on whether it repeats a previous token.
The logit of the repeated token is divided by the penalty before being processed
by the softmax layer.
Our results, shown in \cref{fig:rep}, indicate that a proper repetition penalty
has a positive impact on dialogue state extraction.
The reduction in repetition helps prevent the task bot from mistakenly copying
values directly from the prefix to the suffix, e.g., \exam{Restaurant(name=pizza
  hut,area=pizza hut)}.

\subsubsection{Cause Analysis: Why Longer Context Leads to Lower Precision in
  State Extraction}
\label{sec:why_cont}

In \cref{sec:target_results}, we see that state extraction precision declines as
the length of the input partial dialogue state prefixes increases.
This trend continues until the prefixes reach a certain length (greater than 6
in our analysis), at which point the precision begins to improve again.
This finding contradicts the prior study~\cite{carlini2022quantifying}, which
suggests that an increase in the number of context tokens enhances the ability
to extract memorized text.
To understand why, we explore the unique characteristics of finetuned task bots
considering: 1) how dialogue data repeats in the training dataset and 2) the
one-to-many nature of dialogue, where a single dialogue history can lead to
multiple valid responses.

\textit{Regarding training dialogue data repetition:} When training a task bot
with task-oriented dialogue data, the dialogue is decomposed into turns
consisting of dialogue history, dialogue state, and so on (see
\cref{fig:train_data}).
With this decomposition, the dialogue state that appears early in the dialogue
is repeated more frequently in the training data.
For instance, in a dialogue with $10$ turns, the dialogue state of the first
turn, e.g., \exam{price=cheap}, will also appear in the subsequent
$9$ turns unless the user requests to modify the price range.
Previous studies have shown that repetition can strengthen
memorization~\cite{carlini2021extracting, carlini2022quantifying}.
Therefore, states of greater length, despite having more context, may be harder
to extract.

\begin{figure}[t]
  \centering
  \includegraphics[width=1\linewidth]{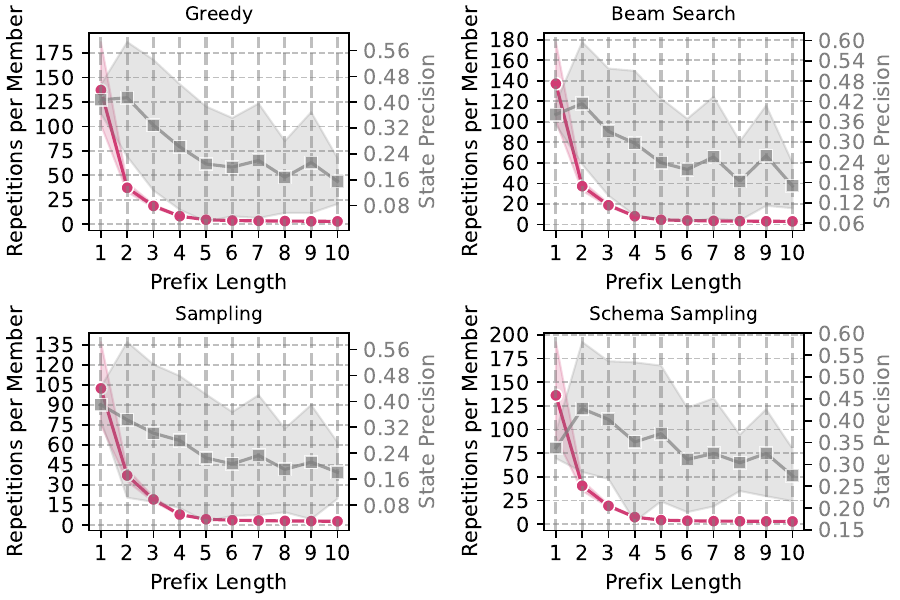}
  \caption{Repetitions per member under different prefix lengths: This measures
    how many times successfully extracted training dialogue states are repeated
    in the training dataset.
    Results are reported under a 99\% confidence interval.
    Additionally, we include the state precision for different extraction
    methods (gray line).}
  \label{fig:freq}
\end{figure}

To verify this, we count how many times the true positive examples we extracted
repeated in the training dataset.
As shown in \cref{fig:freq}, repetitions per member decrease logarithmically
with increasing prefix length.
When the context provided is insufficient, this sharp decline in repetitions can
overwhelm the benefits gained from the increased context, making state
extraction more challenging and leading to decreased state extraction precision.
However, once the context reaches a sufficient length, the advantage shifts
towards the extended context, resulting in a gradual improvement in precision.

\begin{figure}[t]
  \centering
  \includegraphics[width=\linewidth]{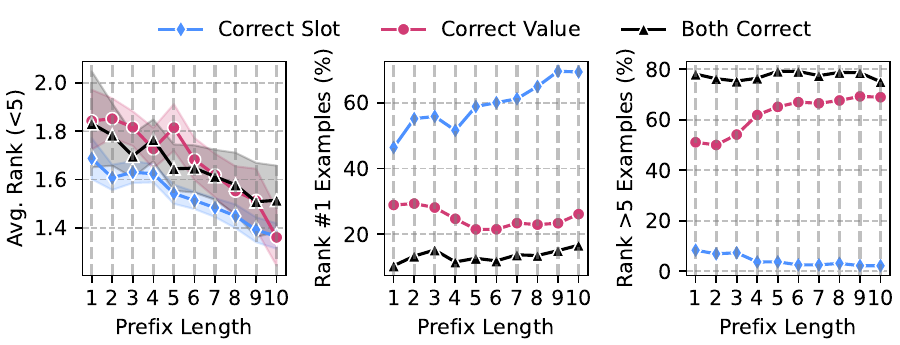}
  \caption{Gold answer ranking based on task bot decoding probability.
    We report results for slots, values, and the combined pairs.
    \textbf{Avg.~Rank (\textless5)} refers to the average rank of examples that
    appear within the top 5.
    A lower rank indicates that the true answer is more confidently decoded by
    the task bot.
    \textbf{Rank \#1} and \textbf{Rank \textgreater 5} represent the percentage
    of true examples that rank at top-1 and those not within the top-5,
    respectively.
    A higher percentage of Rank \#1 and a lower percentage of Rank \textgreater
    5 suggest greater certainty of the task bot in decoding the correct
    answers.}
  \label{fig:logits}
\end{figure}

\begin{table*}[htp]
\centering
\footnotesize
\caption{State and value precision for untargeted dialogue state extraction under various membership inference measures. ``D'' and ``C'' denotes ``Debiased'' and ``conditional'', respectively.}
\label{tab:unt_mia_adv}
\begin{tabular}{lcccccccccc}
\toprule
& \multicolumn{5}{c}{\textbf{State Precision (\%)}} & \multicolumn{5}{c}{\textbf{Value Precision (\%)}} \\
\cmidrule(lr){2-6}
\cmidrule(lr){7-11}
\textbf{Extraction Method} & \textbf{None} & \textbf{PPL} & \textbf{PPL-zlib} & \textbf{C-PPL} & \textbf{DC-PPL} & \textbf{None} & \textbf{PPL} & \textbf{PPL-zlib} & \textbf{C-PPL} & \textbf{DC-PPL} \\
\midrule
Vanilla Sampling & 14.37 & 26.00 & 26.00 & 24.00 & \textbf{42.00} & 15.00 & 60.52 & 54.90 & 55.00 & \textbf{70.27} \\
Temperature Sampling & 4.59 & 19.00 & 11.00 & 14.00 & \textbf{23.00} & 5.76 & 38.56 & 33.00 & 36.95 & \textbf{46.45} \\
Beam Sampling & 11.95 & 23.00 & 21.00 & 29.00 & \textbf{40.00} & 10.73 & 66.97 & 68.10 & 70.19 & \textbf{76.34} \\
\bottomrule
\end{tabular}
\end{table*}

\textit{Regarding the one-to-many feature of task-oriented dialogues:} The state
extraction results in \cref{fig:adv_state_extr} show that while state precision
decreases, value precision concurrently increases.
This suggests that although we can extract values present in the training
dataset, these values do not always constitute a true entailment for the current
prefix.
We hypothesize that this is due to the inherent one-to-many nature of
task-oriented dialogues, where the same context can lead to different valid
responses from different users, thereby associating multiple possible values
with the same prefix.
To investigate this further, we analyzed how the gold slot, gold value, and
their combinations are ranked regarding the decoding probability under different
prefix lengths.
As shown in \cref{fig:logits}, the average ranks decrease as the prefix length
increases.
However, the instances where the gold \emph{Slots} rank first rise
significantly, from $40\%$ to $70\%$, indicating a strong contextual dependency.
In contrast, the percentage of gold \emph{Values} ranking first remains low,
only fluctuating between $20\%$ and $30\%$, which indicates a weaker contextual
dependency.
Despite not frequently ranking first, gold values are increasingly found within
the top five positions, verifying the one-to-many nature that diverse potential
responses can emerge from a single dialogue context.

\subsection{Debiased Membership Inference Results}
\subsubsection{Settings}

We rank the extractions based on several membership inference measures and
retain the top $100$ examples for evaluation.
Please revisit \cref{sec:32} and \cref{sec:adv_mia} for a detailed introduction
to the membership inference methods.

\subsubsection{Results on Targeted Dialogue State Extraction}

\begin{figure}[t]
  \centering
  \includegraphics[width=\linewidth]{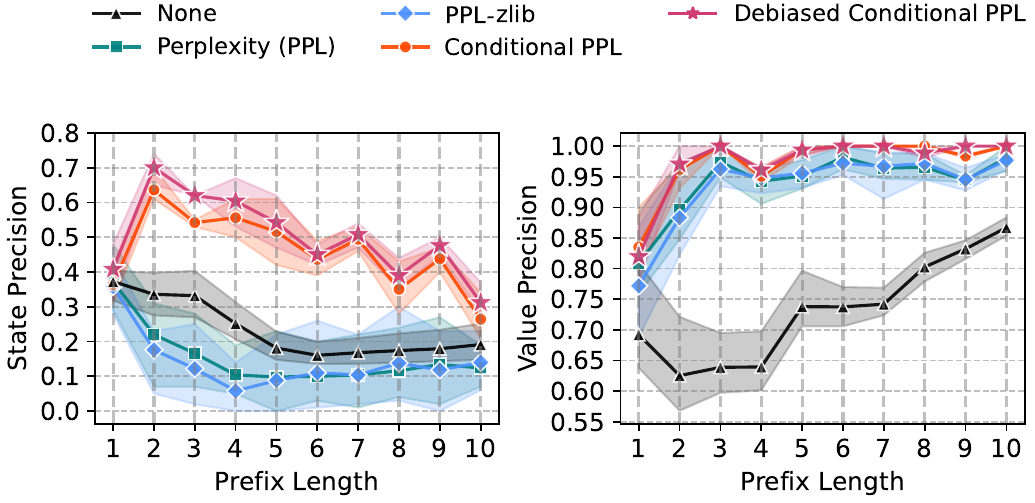}
  \caption{Dialogue state and slot value precision for targeted dialogue state
    extraction after membership inference.
    We report the results averaged
    across various extraction methods detailed in
    \cref{sec:extraction_performance}.
    Shaded areas represent the performance range from the best to the worst
    extraction methods.}
  \label{fig:mia}
\end{figure}

Given the dialogue states extracted using various methods, we apply membership
inference measures to these dialogue states and report the precision of the
final obtained dialogue state and slot value.
As shown in \cref{fig:mia}, we see that both the proposed Conditional PPL and
Debiased Conditional PPL show a marginal improvement in dialogue state precision
and also lead in slot value precision.
Notably, Debiased Conditional PPL performs best, with the best case achieving
over $70\%$ in dialogue state precision and $100\%$ in slot value precision.
We also notice that PPL and PPL-zlib perform worse than random ranking in
dialogue state precision but perform well in value precision.
This discrepancy arises because PPL and PPL-zlib evaluate the full
prefix-suffix sequence.
In targeted extraction, however, the prefixes come directly from the training
data and are thus inherently familiar to the dialogue model.
Consequently, even when paired with wrong suffixes, these familiar prefixes tend
to receive high rankings, introducing a bias that undermines performance.

\begin{figure}[t]
  \centering
  \includegraphics[width=.9\linewidth]{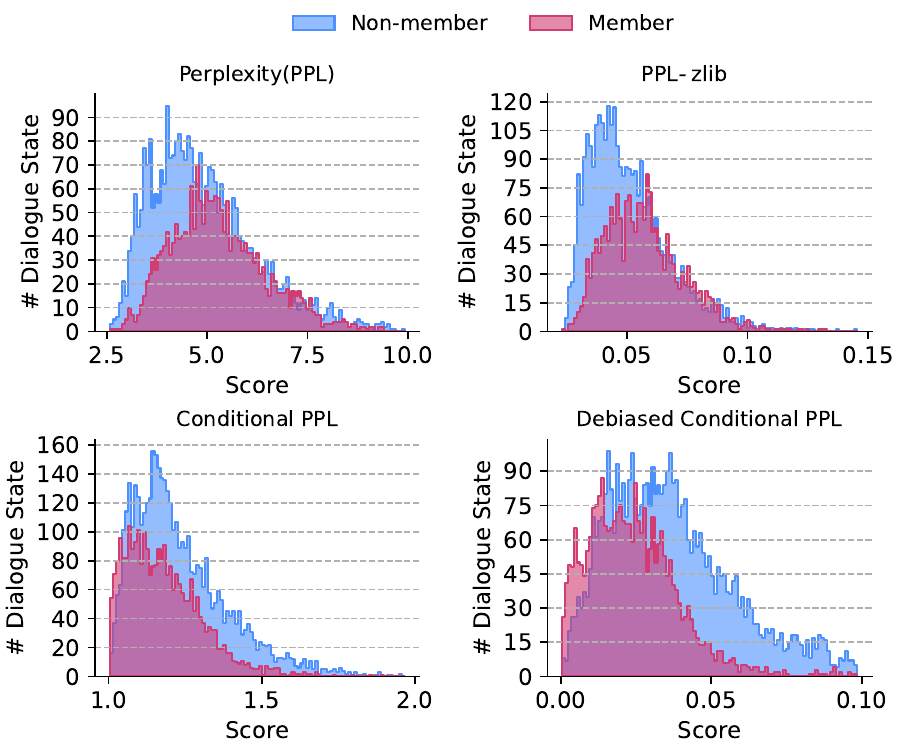}
  \caption{Membership inference score distribution for extracted dialogue
    states.
    We compare dialogue states that appear in the training dataset (Members)
    with those excluded (Non-Members).}
  \label{fig:visual}
\end{figure}

To further study how well different membership inference measures distinguish
extractions, we report score distributions for \emph{ground-truth} members and
non-members.
\cref{fig:visual} reveals that for PPL and ppl-zlib, the distributions
completely overlap, making it impossible to distinguish members and non-members.
In contrast, although Conditional PPL and Debiased Conditional PPL show some
overlap, members can be flagged in the lower score range, demonstrating the
effectiveness.

\subsubsection{Results on Untargeted Dialogue State Extraction}
Targeted extraction employs partial gold prefixes
from the training dataset, potentially disadvantaging PPL and PPL-zlib.
Therefore, we also report the results under an untargeted extraction attack
setting, where no partial dialogue state is included.
In this setting, the context for Conditional PPL and its variant is simply
\exam{Belief State:} as described in \cref{sec:31}.

As shown in \cref{tab:unt_mia_adv}, our proposed Debiased Conditional PPL
continues to outperform baseline methods significantly in both dialogue state
and slot value precision, demonstrating its effectiveness in detecting training
dialogue states.
However, Conditional PPL does not perform as well, as without a partial dialogue
state, it exaggerates the inherent bias of perplexity towards repetitive or
longer sequences.

\begin{figure}[t]
  \centering
  \subfloat[]{%
    \includegraphics[width=.5\linewidth]{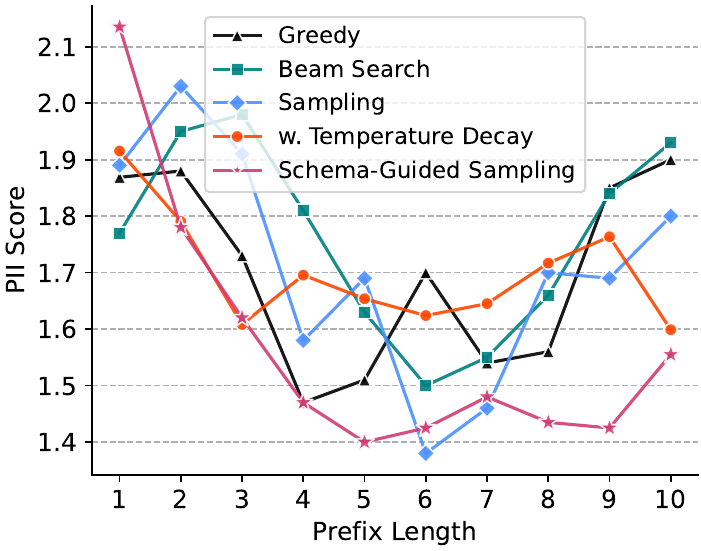}}
  \subfloat[]{%
    \includegraphics[width=.5\linewidth]{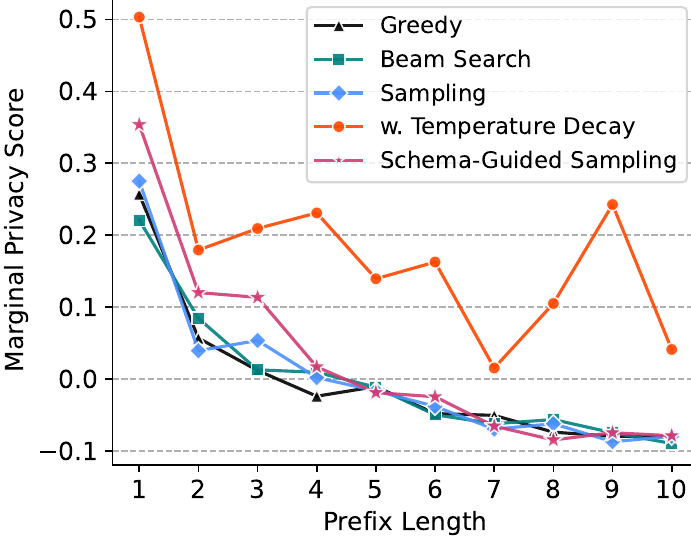}}\\
  \subfloat[]{%
    \includegraphics[width=.5\linewidth]{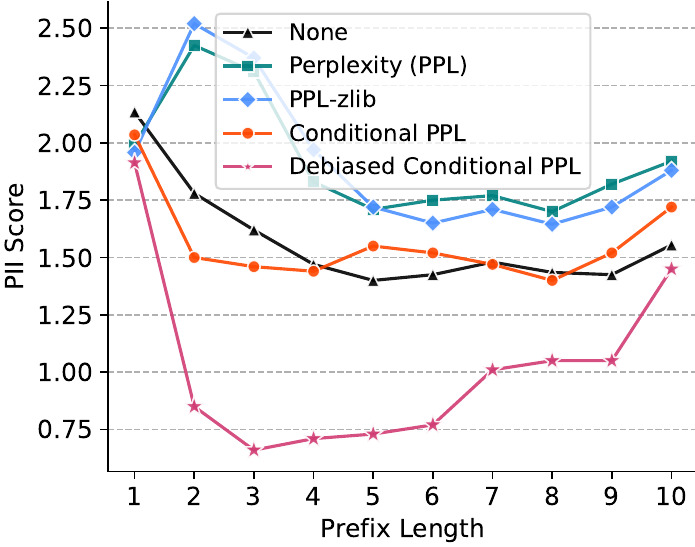}}
  \subfloat[]{%
    \includegraphics[width=.5\linewidth]{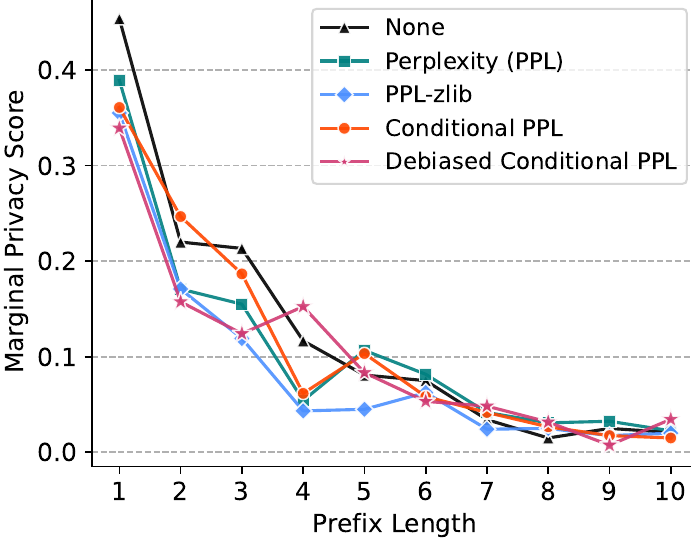}}\\
  \caption{PII score and marginal privacy score under different training data extraction (a) (b) and membership inference methods (c) (d).}
  \label{fig:privacy}
\end{figure}

\subsection{Privacy Level Analysis Results}
{To quantitatively assess the privacy sensitivity of the extracted information, we conducted a fine-grained privacy level analysis. This experiment was designed to validate our core argument that privacy risk arises not only from direct PII (e.g., names, phone numbers) but also from combinational PII, where seemingly innocuous data points can reveal sensitive information when combined. To this end, we employed the prompt-based deepseek-rl-671B model as an expert assessor to evaluate the privacy degree of each extracted belief state (event-level) against a strict protocol. The model assigned an overall privacy score from 0 (no concern) to 5 (very high), adhering to a critical monotonicity rule: a longer belief state (superset) could not receive a lower score than any of its subsets, ensuring that added information never reduces perceived sensitivity.
}

{Two specific metrics are computed:}
\begin{itemize}
    
    \item {\textbf{PII Score:} The privacy score for a single \textit{suffix}, obtained by first scoring each of its constituent \textit{domain-slot-value triplets} individually and then averaging these scores. This metric reflects the inherent sensitivity of the suffix when its components are viewed in isolation.}
    \item {\textbf{Marginal Privacy Score:} This metric quantifies the relative increase in privacy risk introduced by incorporating the extracted suffix into the existing context. It is calculated as the ratio of the \textit{increase} in privacy score to the original score of the prefix, formally as:}
\[
{
\text{Marginal Privacy Score} = \frac{S_{\text{prefix + suffix}} - S_{\text{prefix}}}{S_{\text{prefix}}}
}
\]
{where \( S_{\text{prefix}} \) and \( S_{\text{prefix + suffix}} \) are the overall privacy scores of the prefix and the combined state, respectively. A higher value indicates a greater amplification of privacy sensitivity, thereby pinpointing instances where combinational PII is created.}
\end{itemize}

{As shown in \cref{fig:privacy}, the results reveal a notable distinction between the extraction methods. Traditional decoding and membership inference methods generally yielded suffixes with higher individual PII scores, indicating a tendency to extract locally sensitive information in isolation. In contrast, our schema sampling and debiasd membership inference methods, while not achieving the highest individual PII scores, achieved comparable or even higher marginal privacy scores. 
We further verified the reliability of the automated assessor via human evaluation, and observed substantial agreement between model scores and human judgments, with geometric mean Cohen’s $\kappa$ of $0.858$ (Marginal Privacy Score) and $0.761$ (PII Score).
This finding is significant because it demonstrates that our approach is more effective at uncovering meaningful event-level information from a holistic context, precisely by revealing combinational PII. Crucially, this superior capability to expose the privacy risks arising from data combinations is achieved while our method extracts a substantially larger number of valid events with much higher membership inference accuracy, thereby establishing a more conservative lower bound on potential privacy risks.
}

\section{Towards Defense and Protecting Privacy}
\label{sec:def}




As discussed in \cref{sec:why_cont}, we identify two issues significantly
affecting extraction results.
1) Data repetition from turn-level modeling, i.e., the dialogue state in the
first turn appears in all subsequent turns, enhances memorization.
2) The one-to-many nature of dialogue, where a given dialogue history might
yield multiple valid user choices, leading to no single certain gold answer,
reduces memorization.
Given these features, we propose two potential privacy-enhancing methods:

\emph{1) Dialogue-Level Modeling:} To reduce data repetition associated with
turn-level modeling, we could model dialogue at the dialogue level, passing the
entire dialogue per training instance.
Specifically, given a full dialogue consisting of multiple turns, $H_t =
\{U_0,R_0,\ldots,U_{t-1},R_{t-1},U_t\}$, which includes both user utterances and
previous system responses.
The current training methodology utilizes turn-level data, represented as
$\{U_0, R_0, \ldots, R_{t-1}, U_t, S_t, D_t, A_t, R_t\}$, where $t$ ranges from
$1$ to $T$.
We propose to transform it into a dialogue-level model input: $\{U_0, S'_0, D_0,
A_0, R_0, \ldots, U_T, S'_T, D_T, A_T, R_T\}$.
Here, $S'$ captures the dialogue state of the current turn rather than the full
dialogue history, thus preventing previous turn dialogue states from appearing
multiple times.
Training in this manner involves conditional language modeling and loss
calculation on the dialogue states, actions, and responses to update the model.
Note that while this reduces repetition, it also demands more GPU memory and
more data.

{\emph{2) Value Copy Mechanism:} 
Given the one-to-many nature of task-oriented dialogues, a more privacy-preserving variant of task bots could adopt a value copy mechanism, in which slot values are not newly generated from contextual modeling but are directly copied from the dialogue history. 
Under this mechanism, without the history, value generation would produce a blank, thus preventing unintended reproduction of memorized values. 
}

{Practically, such behavior can be encouraged through counterfactual data augmentation that randomly replaces or removes slot values in both the dialogue history and the corresponding belief states during training.
This procedure diversifies value distributions and weakens the model’s dependence on specific values while keeping the semantic structure of dialogues intact.
Since task-oriented dialogue performance mainly depends on accurate schema tracking and turn-level reasoning rather than memorization of concrete slot values, this augmentation regularizes value generation without causing significant performance degradation.
In essence, the approach improves robustness and privacy safety by encouraging the model to generalize over slot patterns instead of memorizing specific user-provided values.
}

\section{Conclusion} \label{sec:conclusion}

This work explores how LLMs memorize and regenerate training data when finetuned
for the downstream application of task-oriented dialogue systems, i.e., task
bots.
By performing training data extraction attacks, we first show that LLM-based task bots \emph{do} memorize and
inadvertently leak training labels of dialogue states even without conditioning
on any context.
Our results show that values such as phone numbers are easier to extract, with maximum
precision reaching $67\%$, and the full dialogue states present a significantly
lesser concern, with a maximum precision of only $26\%$.
We further investigate the targeted extraction setting where we seed the task
bot with partial dialogue states as conditioning context to extract the rest,
and make several improvements regarding suffix decoding and membership
inference.
Empirical results showed a significant improvement, reaching a best-case
scenario of $100\%$ extraction precision for values and a maximum of over $70\%$
for dialogue states.
We also investigate the factors influencing memorization and pose several
potential mitigating strategies to guide further development of LLM-based task bots.

\section*{Acknowledgment}
This work was supported in part by National Key Research and Development Plan in China (2023YFC3306100) and National Natural Science Foundation of China (62272372).



\bibliographystyle{IEEEtran}
\bibliography{acl2020}

\end{document}